\documentclass[twoside,leqno,twocolumn]{article}
\usepackage{ltexpprt}

\usepackage{array}
\usepackage{amsmath}
\usepackage{amsfonts}
\usepackage{amssymb}
\usepackage{algorithm}
\usepackage{algorithmicx}
\usepackage[noend]{algpseudocode}
\usepackage{algpseudocode}
\usepackage{adjustbox}

\usepackage{boldline}
\usepackage[english]{babel}
\usepackage{booktabs}

\usepackage{comment}
\usepackage{color}

\usepackage{diagbox}
\usepackage{dirtytalk}

\usepackage{enumitem}

\usepackage{graphicx}
\usepackage{hyperref}
\usepackage[utf8]{inputenc}

\usepackage{mathptmx}
\usepackage{multirow}
\usepackage{makecell}

\usepackage{pifont}

\usepackage{soul}
\usepackage{setspace}
\usepackage{subcaption}

\usepackage{tabu}
\usepackage[para]{threeparttable} 

\usepackage{xfrac}
\usepackage{xspace}

\algrenewcommand\algorithmicindent{1.0em}

\newcommand{\MF}{\mbox{MF}\xspace}
\newcommand{\MFwithbias}{\mbox{MF-b}\xspace}
\newcommand{\MFnobias}{\mbox{MF}\xspace}
\newcommand{\DomainAware}{\mbox{MF-d}\xspace}
\newcommand{\CKRM}{\mbox{CKRM}\xspace}
\newcommand{\AvgCKRM}{\mbox{ACK}\xspace}
\newcommand{\AvgCKRMb}{\mbox{ACK-b}\xspace}

\newcommand{\EnMF}{\mbox{NAN}\xspace}

\newcommand{\EnMFb}{\mbox{NAN}\xspace}
\newcommand{\EnFARM}{\mbox{NAN}\xspace}
\newcommand{\EnFARMb}{\mbox{NAN}\xspace}

\newcommand{\EnFARMa}{\mbox{ALE}\xspace}
\newcommand{\EnFARMain}{\mbox{ALE$_{\neg in}$}\xspace}
\newcommand{\EnFARMaal}{\mbox{ALE$_{\neg al}$}\xspace}
\newcommand{\EnFARMaI}{\mbox{ALE$_{\neg g}$}\xspace}
\newcommand{\EnFARMab}{\mbox{ALE-b}\xspace}
\newcommand{\PTAzero}{\mbox{PTA$_0$}\xspace}
\newcommand{\PTA}{\mbox{PTA}\xspace}
\newcommand{\PTAone}{\mbox{PTA$_1$}\xspace}
\newcommand{\PTAtwo}{\mbox{PTA$_2$}\xspace}
\newcommand{\tickNo}{\text{\ding{55}}}
\newcommand{\method}{\mbox{ALE}\xspace}

\begin{document}

\title{\Large ALE: Additive Latent Effect Models for Grade Prediction}

\author{Zhiyun Ren\thanks{Department of Computer Science, George Mason University.}
\and
Xia Ning\thanks{Department of Computer and Information Science, Indiana University-Purdue University Indianapolis.}
\and
Huzefa Rangwala\footnotemark[1]
}
\date{}

\maketitle


\fancyfoot[R]{\footnotesize{\textbf{Copyright \textcopyright\ 20XX by SIAM\\
Unauthorized reproduction of this article is prohibited}}}





\begin{abstract} 
The past decade has seen a growth 
in the development 
and deployment of educational technologies 
for assisting college-going students  in choosing majors, selecting 
  courses and acquiring
  feedback based on past academic performance. 
Grade 
  prediction 
  methods seek to estimate a grade
  that a student may achieve in a
  course that she may take in the  future (e.g., next term). Accurate and timely prediction 
  of students' academic grades  is important for developing  effective 
  degree planners and early warning systems, and ultimately improving educational
  outcomes. 
 Existing grade prediction methods mostly focus on modeling the knowledge components associated with each course and student, and often overlook other factors such as the difficulty of each knowledge component, course instructors, student interest, capabilities and effort.

In this paper, we propose additive latent effect models that incorporate these factors to predict the student next-term grades.
Specifically, the proposed models take into account four factors: (i) student's academic level, (ii) course instructors, (iii) student global latent factor, 
and (iv) latent knowledge factors. 
We compared the new models with several state-of-the-art methods on students of various characteristics (e.g., whether a student transferred in or not). The experimental results demonstrate that the proposed methods significantly outperform the 
baselines on grade prediction problem.
Moreover, we perform a  
thorough analysis on the importance of different factors and how these factors can practically assist students in course selection, and finally improve their academic performance.

\end{abstract}

\section{Introduction.}

One of the grand challenges facing higher education institutions, (i.e., 
four-year colleges/universities and community colleges) is 
low graduation rates \cite{parker2015advising}. To increase student graduation rates, several 
educational data mining 
techniques have been developed and deployed at
several institutions 
to provide students degree pathways towards successful graduation  ~\cite{simons2011national}. Additionally, 
early warning systems have been developed to monitor student progress, and identify students at-risk of dropping majors or performing 
below their potential in a given major. 
An effective way to assist and improve degree planning and advising is via modeling the student's knowledge and foreseeing their future academic performance \cite{parker2015advising}.

Matrix factorization (\MF) based approaches 
have been widely used for solving the grade prediction problems \cite{Mack2, elbadrawy2016predicting}.
MF 
methods decompose the student-course grade matrix into two 
low-rank matrices containing student and course latent factors. The prediction of a 
student's grade on an untaken course will be calculated as
the similarity of the corresponding student latent factors and course latent factors. However, the existing 
grade prediction methods often have a narrow focus on the potential influential factors. For example, course instructors, course difficulty, student's interest, capability and effort are rarely considered
in the existing methods, which are all  important factors to the student's grades.

\begin{figure}[!]
         \begin{subfigure}[b]{0.4\linewidth}
                 \centering
                 \includegraphics[width=0.98\textwidth]{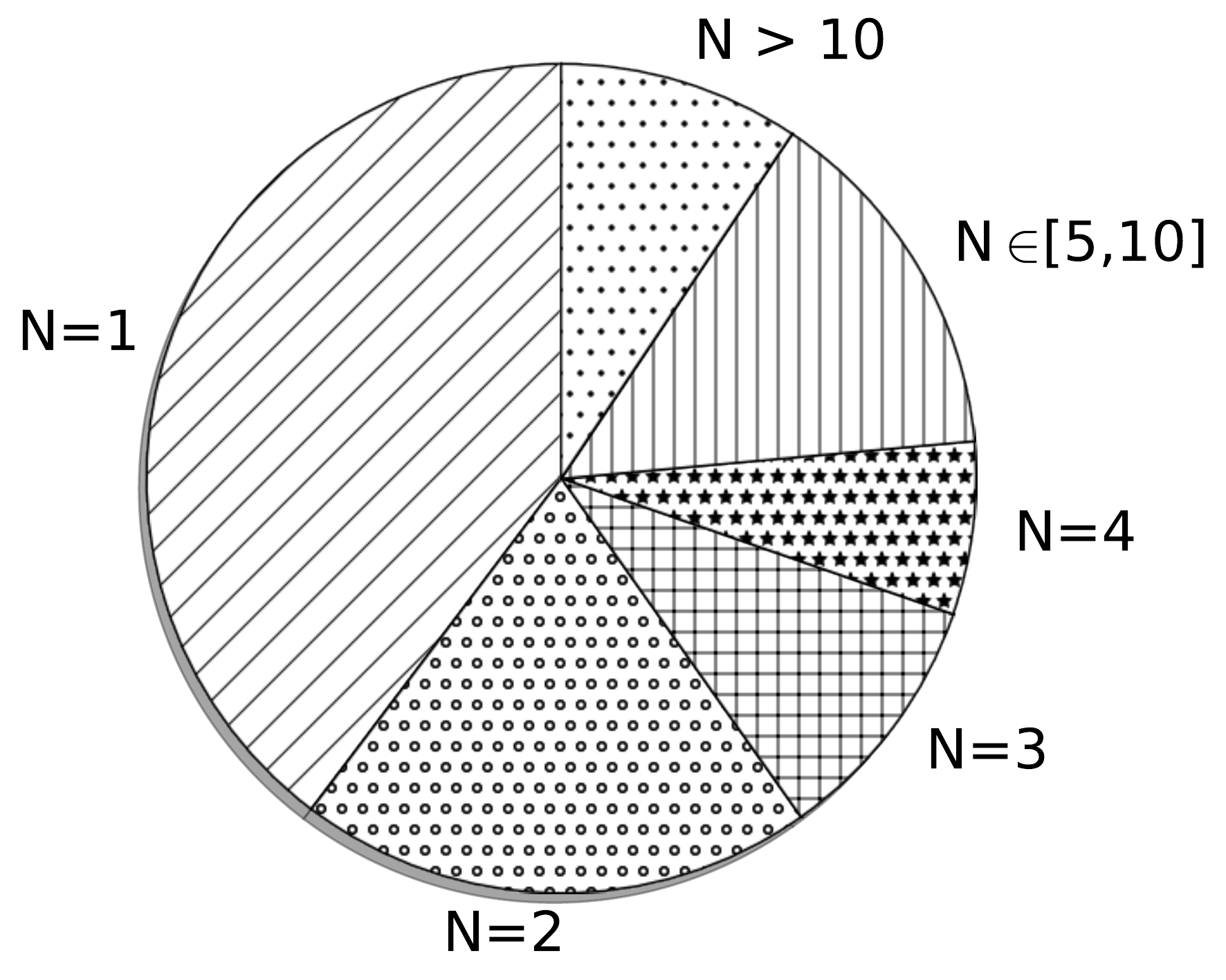}
                 \caption{Distribution of \#instructors over courses}
                 \label{fig:crsPie}
         \end{subfigure}\hfill
         \begin{subfigure}[b]{0.57\linewidth}
                 \centering
                 \includegraphics[width=0.98\textwidth]{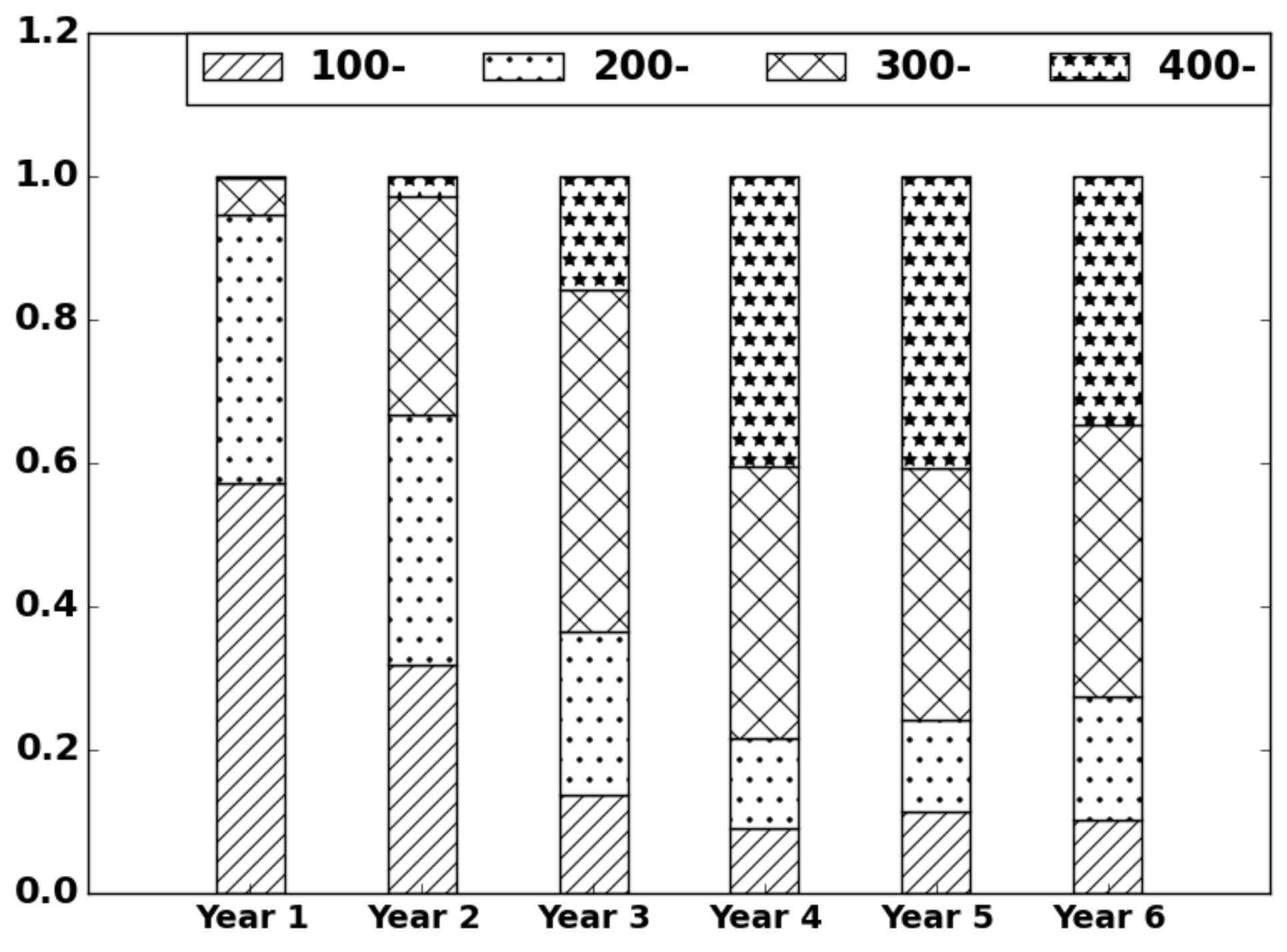}
                 \caption{ \% of Students Enrolled in Different Course Levels based on their Current College Year}
                 \label{fig:stdBar}
         \end{subfigure}%
         \caption{Course-Student Data Distributions}~\label{fig:distribution}\
\end{figure}

\textit{Data distribution:}
 Fig.~\ref{fig:crsPie} shows 
the distribution of the number of 
instructors who teach
the same course  at George Mason University. More than 60\% 
of the courses at this university have been taught by multiple instructors in a period of 18 terms. 
For a given course, different instructors differ in their course offerings with respect to coverage of 
 course topics, pedagogy and grading criterion.  All these factors impact a 
 student's grade in a 
 course.
  As such, 
 we propose to model latent factors associated with each instructor in addition to the latent factors of the course she teaches.
  %
  Fig.~\ref{fig:stdBar} shows 
the distribution 
of academic course levels at George Mason University (i.e., 100-,200-,300- and 400-level)
offered to the students in 
different starting years.
%
We assume  that students in the 
same  college terms (e.g., freshmen, sophomore, etc)  
tend to have similar learning behaviors, capabilities and expertise   
given the 
sequential aspects of most degree programs.
For example, freshmen students may be undecided
on 
their 
majors and mostly take courses with level 100, as shown in Fig.\ref{fig:stdBar}.
Likewise, seniors tend to have an 
in-depth knowledge of  study in a specific field, and mostly take higher level courses.

In this work, we propose {\bf A}dditive {\bf L}atent {\bf E}ffect (\textbf{\method}) models within the framework of MF
to predict the grade that a student is expected to obtain in a course that she may enroll in the next term.
Inspired by Morsy et. al. \cite{morsy2017cumulative}, the proposed 
methods model each student's latent factors with accumulated knowledge of a sequence of courses taken by the student, jointly with the grade for each course. Furthermore,  
we incorporate 
course  instructor and 
student academic level effects along with 
student global latent factor  to 
enable accurate grade prediction. %

 We conducted a comprehensive set of experiments on various datasets 
and
provided  a  thorough analysis on the importance of different factors.    
Our experimental results  show that the proposed methods achieve 
superior prediction performance 
on various test datasets for 
 next-term grade prediction.

The main contributions of our work in this paper can be summarized as follows:
\begin{enumerate}
\item  We propose additive latent effect models that 
incorporate 
the information of 
course instructors, student's academic level and student global latent factor  
for the next-term grade prediction problem.   The strengths of our proposed framework include the ability 
to enhance the standard \MF methods with additional student and course-specific content 
information that may not be contained within the student-course grade matrix. 
\item We implement a number of extensive experiments on different student groups partitioned by student starting years and student majors. We then provide detailed analysis on the importance of each additive effect in our model, and how to assist students in selecting courses.

\end{enumerate}

\section{Related Work.}
Over the past few years, a large number of methods  influenced by Recommender System (RS) research \cite{Aggarwal2016}, including 
Collaborative Filtering (CF) \cite{Ning2015} and Matrix Factorization \cite{Koren2009}
have inspired the development of
methods for educational data mining to solve the next-term 
grade prediction problem \cite{thai2010recommender} and in-class grade prediction problem \cite{elbadrawy2014personalized}. 
%
For example, Sahebi \emph{et. al.} \cite{sahebi2016tensor} modeled student
learning progress and predicted student performance using tensor
decomposition based on the sequence of student attempts within course quizzes. 
Lan \emph{et. al.} \cite{lan2016dealbreaker} predicted student performance 
on different questions 
within the context of intelligent tutoring systems.
Meier \emph{et. al.} \cite{meier2015personalized} developed an 
online learning method that learns  the best 
time to intervene based on past student performance in
a course. 
Additionally, incorporation of biases has shown 
to be important for several
educational data mining problems, following
its success in RS \cite{Koren2009}. 
Elbadrawy \emph{et. al.} \cite{elbadrawy2016domain}  developed 
a domain-aware grade prediction method with student/course-group based biases.
To predict student $s$' grade on course $c$, this 
method groups students and courses in different ways 
based on student majors, academic levels and course subjects, and
introduces group-based biases.
The  key intuition of this method is 
that 
students who take the same course and can be grouped by domain information (e.g., student's major) 
may share a similar bias.


In educational  data mining problems, 
sequential information of students/courses over time
is very common and thus methods that deal with sequential data can be
beneficial. As a matter of fact, such methods have been extensively developed
in RS research. For example, 
%
integrated methods of Markov Chains (MC) and \MF have been popular in dealing with sequential data in RS. 
Rendle \emph{et. al.}  \cite{rendle2010factorizing} proposed the 
factorized personalized MC models. These models have personalized Markov chains 
that rely on
transition matrices, and these methods use a factorization model to deal with the 
sparsity in the input data. 
He \emph{et. al.} \cite{he2016fusing} developed 
factorized sequential models with item similarities for sparse 
sequential recommendation. Their models consider both long-term and short-term dynamics
among user-item data.
%
He \emph{et. al.} \cite{he2016vista} adopted a 
similar idea and developed large-scale recommender systems to model
the 
preferences and short-term dynamics between both users and items.



%

Prior work in the  RS literature 
that shares similarities with our proposed method is from
Koenigstein \emph{et. al.} \cite{koenigstein2011yahoo}.
In this work, the authors 
proposed a 
music  rating recommendation system  that models a user's music preferences 
based on her interest in a given music track, and
the artist and album information associated with the specific track.
Shared factor components were introduced to 
reflect the similar preference for music tracks of same artists (or genre, album).  
%
%

We will discuss in detail the
domain-aware grade prediction method \cite{elbadrawy2016domain} and 
cumulative knowledge-based regression
model  \cite{morsy2017cumulative} in Section {\ref{sec:background}} as they serve as
foundations for our proposed formulation. 



\section{Notations and Definitions}

Formally, student-course grades will be represented by 
\{$G_1$, $G_2$, ..., $G_N$\}  for $N$ terms. 
$G_t$ contains the set of  
tuples storing grade information 
for all students enrolled in courses within  term $t$. Each tuple 
  stores: (i) student identifier, (ii) course identifier, (iii) student academic level, (iv) course instructor, and 
  (v) grade obtained.  
For all students, the student-course 
grades 
up to the term $t$ can be 
represented
by $G^t$=$\bigcup_{i=1}^{t}$$G_i$.
  The set of courses that student $s$ has taken 
  in term $t$ is represented by $C_{s,t}$ and 
the grades that student $s$ achieves 
in term $t$ is
  represented by $G_{s, t}$.
 The  set of courses that student $s$ has taken up to term $t$ is represented by $C_s^t$, 
 and  the grades 
  that student $s$ has achieved 
  up to term $t$ is represented
  by $G_s^t$.

In this paper, all vectors (e.g., $\mathbf{p}_s^{\mathsf{T}}$ and $\mathbf{q}_c$) are represented by bold lower-case letters and all matrices (e.g., $A$) are represented 
by upper-case letters.  
Row vectors are represented by having the transpose superscript$^{\mathsf{T}}$, otherwise by default they are 
column vectors. A predicted value is denoted by having a $\tilde \ $ symbol. Table \ref{tab:Notation} summarizes the key notations used in this paper.

Given student-course grades up to term $(T-1)$,  
the objective of 
the next-term grade prediction problem 
is to predict grades for each student on courses   
that the student may consider for enrollment  in the next term $T$.

\begin{table}
\caption{Notations}~\label{tab:Notation}
\centering
\begin{tabular}{
                @{\hspace{0pt}}>{\small}l@{\hspace{2pt}}
                @{\hspace{2pt}}>{\small}l@{\hspace{0pt}}}
\hline\hline
Notation  & Explanation \\
\hline
$m$ 			&  number of courses \\
$n$ 			&  number of students \\
$k$ 			&  the dimension of latent factors\\
\hline
$\mathbf{p}_{ck(s)}^t$ 			&  latent factors of accumulated knowledge  of student $s$\\
							& up to term $t$\\
$\mathbf{p}_{al(s)}$ 			& latent factor of student $s$'s academic level, \\
								& $al(s) \in [1,12]$ \\
$\mathbf{p}_{en(s)}^t$ 			& integrated student latent factor\\
$\mathbf{q}_c$ 			& latent factor of the knowledge components required\\
					& by course $c$ \\
$\mathbf{q}_{in(c)}$ 			&  latent factor of the instructor who teaches course $c$ \\
$\mathbf{q}_{en{(c)}}$		& integrated course latent factor\\
$\mathbf{p}_{g(s)}$ 			&  student $s$'s global latent factor\\
$\mathbf{k}_c$ 			& latent factor of the knowledge components provided\\
					& by course $c$ \\
\hline
$G_t$ 		&   student-course grades at term $t$ \\
$G^t$ 		&   all the student-course grades up to term $t$  \\
$g_{s,c}^t$ 	&   the grade of student $s$ on course $c$ at term $t$\\
$C_{s,t}$ 		&   the set of courses student $s$ chooses at term $t$\\
$C_s^t$ 		&   the set of courses student $s$ chooses up to term $t$\\
$G_{s,t}$ 		&   all the grades student $s$ obtains at term $t$   \\
$G_s^t$ 		&   all the grades student $s$ obtains up to term $t$ \\
$t_{s,c}$ 		&  the academic term when student $s$ takes course $c$ \\
\hline\hline
\end{tabular}

\end{table}

\section{Background and Prior Methods}
\label{sec:background}

\subsection{Matrix Factorization Based Grade Prediction}
Matrix factorization from RS~\cite{ricci2011recommender}  
 can be
  applied for the next-term grade prediction problem, when the student-course grade matrix is considered
  as the user-item rating matrix.
Two low-rank matrices containing latent
  factors of courses and students in a common knowledge space 
  can be learned from such a
student-course grade matrix ~\cite{Mack2}.
Thus, the grade of a student $s$ on a course $c$ can be predicted as
 \begin{eqnarray}
   \label{eq:mf0}
   \begin{aligned}
     \tilde g_{s,c}  = \mathbf{p}_s^{\mathsf{T}} \mathbf{q}_c, 
   \end{aligned}
 \end{eqnarray}
 where $\mathbf{p}_s$ ($\mathbf{p}_s \in \mathbb{R}^{k}$)
 and $\mathbf{q}_c$ ($\mathbf{q}_c \in \mathbb{R}^{k}$)
 are the two vectors containing latent factors of $k$ dimensions for student $s$ and course $c$, respectively.  
   This method is denoted as \MF.

Including the bias terms within the \MF formulation 
has shown to be 
   effective in modeling systematical biases~\cite{Koren2009}. 
 For the grade prediction problem 
 using \MF, student and course biases
 can be included as follows:
 \begin{eqnarray}
   \label{eq:mf}
   \begin{aligned}
     \tilde g_{s,c}  =  \mathbf{p}_s^{\mathsf{T}} \mathbf{q}_c + b_s + b_c, 
   \end{aligned}
 \end{eqnarray}
 where $b_s$ and $b_c$ are bias terms for student $s$ and course $c$, respectively.
 This method is
 referred to as \MF with bias terms and denoted as \MFwithbias.

\subsection{\MF with Domain-Aware Biases}

 El-Badrawy \textit{et. al.} developed a domain-aware \MF based methods for the
   next-term grade prediction problem~\cite{elbadrawy2016domain}. These methods involve 
    group-specific bias within  \MF formulation. Groups are defined 
   based on student- and course-specific information.

In this method, the grade for  student $s$ on course $c$ is predicted as:
 \begin{eqnarray}
   \label{eq:mfd}
   \begin{aligned}
     \tilde g_{s,c}  =  \mathbf{p}_s^{\mathsf{T}} \mathbf{q}_c + b_s^{\varphi(c)} + b_c^{\varphi(s)},
   \end{aligned}
 \end{eqnarray}
 where $\mathbf{p}_s$ ($\mathbf{p}_s \in \mathbb{R}^{k}$)
 and $\mathbf{q}_c$ ($\mathbf{q}_c \in \mathbb{R}^{k}$)
 are the latent factors for student $s$ and course $c$, respectively. 
 $\varphi()$ denotes the grouping information. 
$b_c^{\varphi(s)}$ is the bias term for course $c$.
This method models course bias  based on the performance of 
students who are in the same group of student $s$ (i.e., $\varphi(s)$) and have taken course $c$ before.
Similarly, $b_s^{\varphi(c)}$ is the student bias term modeled based on the grades student $s$ has got on the
courses which are in the same group as course $c$ (i.e., $\varphi(c)$).
%
%
%
%
 We refer to this method 
as \MF with domain-aware biases and denote it as \DomainAware.


 \subsection{Cumulative Knowledge-based Regression Models}
Morsy et. al. \cite{morsy2017cumulative} consider the series of courses a student takes as a sequence and propose 
Cumulative-Knowledge Regression Models (CKRM). 
Specifically, to predict student $s$'s performance on course $c$, CKRM represents student $s$ with the series of courses she has taken in the past, and each course is represented by a vector 
which is expected to capture the latent knowledge components provided by the course. 
Moreover, CKRM represents course $c$ with a vector which is expected to capture the latent  knowledge components required by the course.
Consequently, given a student $s$ in term $t$, $\mathbf{p}_{ck(s)}^t$ is 
the cumulative knowledge acquired until term $t$, and is given by: 

 \begin{equation}
   \label{eq:ck}
   \begin{aligned}
     \mathbf{p}_{ck(s)}^t  =   \sum_{g_{s,c'} \in G_s^{t-1}}(e^{-\lambda(t-t_{s,c'})} \mathbf{k}_{c'} \cdot g_{s,c'}),
   \end{aligned}
 \end{equation}
 where 
$t_{s, c'}$ is the term in which student $s$ took course $c'$,
   $e^{-\lambda(t-t_{s,c'})}$ is an exponential time decay function,
$\mathbf{k}_{c'}$ contains the latent knowledge factors of course $c'$,
 and $g_{s, c'}$ is the grade of
 student $s$ on course $c'$.
 The 
   grade of student $s$ on course $c$ is then predicted as follows:
 \begin{equation}
   \label{eq:ckrm}
   \begin{aligned}
     \tilde g_{s,c}^t  =  {\mathbf{p}_{ck(s)}^t}^{\mathsf{T}} \mathbf{q}_c. 
   \end{aligned}
 \end{equation}
 In this study, we average the results  
 in Eq~ \ref{eq:ckrm} with the 
sum of exponential time decay weight,
 that is: 
\begin{equation}
  \label{eq:AvgCKRM}
  \begin{aligned}
     \tilde g_{s,c}^t  =   \frac{1}{|G_s^{t-1}|} {\mathbf{p}_{ck(s)}^t}^{\mathsf{T}}  \mathbf{q}_c
  \end{aligned}
\end{equation}
%
This method is referred to as \textbf{A}veraged \textbf{CK}RM and denoted as \AvgCKRM. Our preliminary experiments
demonstrate that \AvgCKRM outperforms \CKRM. 

\section{Additive Latent Effect Models (\EnFARMa)}
We propose Additive Latent Effect (\EnFARMa) models  
to predict 
student $s$'s performance on course $c$  in
term $t$.
We will give a thorough presentation on how we model each effect in the following sections.

\subsection{Student Academic Level Effect}
Based on our  assumption that students on a same academic level (i.e., freshmen, sophomore, junior and senior)  have a similar level of academic maturity, experience, habits and knowledge,  
  we model student $s$ by  
    integrating a factor associated to the college term she is attending,
    denoted as $\mathbf{p}_{al(s)^t}$,
    into the student accumulated knowledge factors. The integrated student latent
     factor is  denoted as
    $\mathbf{p}_{en(s)}^t$, and is given as follows: 
  \begin{eqnarray}
    \label{eq:pswithst}
    \begin{aligned}
      \mathbf{p}_{en(s)}^t  = \mathbf{p}_{ck(s)}^t +\mathbf{p}_{al(s)^t},
    \end{aligned}
  \end{eqnarray}
    where $\mathbf{p}_{ck(s)}^t$ is calculated by Eq \ref{eq:AvgCKRM}, and  $al(s)^t$ represents the academic level
    of student $s$ in term $t$, defined as $al(s)^t = t - (\text{$s$'s start term})$.
  Since most students finish college in 4-6 years (8-12 terms),
   $al(s)^t$ is in $[0, 12)$.
We include $\ell_1$-norm regularization on $\mathbf{p}_{al(s)^t}$ to enforce sparsity on this representation. 
This is because  $\mathbf{p}_{al(s)^t}$ aims to capture the academic factors (e.g., academic maturity), and 
student $s$ is only able to hold a part of them on a particular academic level (e.g., student $s$ cannot be both mature and  immature at the same time).

\subsection{Course Instructor Effect}
Consider that a single course is often taught by multiple instructors who usually vary in their coverage of materials (topics), pedagogy, use of teaching technology, choice of assignments and grading criterion.  
We hypothesize that a student's performance on a specific course is greatly influenced by the  
instructor who teaches her the course. 
Specifically,  for a course $c$, we  add a factor associated with the specific instructor who teaches course $c$, denoted as $\mathbf{q}_{in(c)}$, to the original knowledge latent factors of course $c$.
The integrated course latent factor
is denoted by $\mathbf{q}_{en{(c)}}$, and is given as follows:
  \begin{eqnarray}
    \label{eq:qcwithin}
    \begin{aligned}
      \mathbf{q}_{en(c)}  = \mathbf{q}_c +\mathbf{q}_{in(c)},
    \end{aligned}
  \end{eqnarray}
where  $in(c)$ denotes the instructor who teaches 
course $c$. 
 For $\mathbf{q}_{in(c)}$,  we include $\ell_1$-norm regularization
  to control its sparsity. We assume that an instructor is generally  proficient only in certain
    topics (and knowledge components), but not all.

With the course
instructor information and 
student academic level information as proposed above, the 
grade prediction for student $s$ on course $c$  
 is given 
 as follows: 
\begin{eqnarray}
  \label{eq:EnMF}
  \begin{aligned}
\tilde g_{s,c}^t   = {\mathbf{p}_{en(s)}}^{\mathsf{T}} \mathbf{q}_{en(c)}.
  \end{aligned}
 \end{eqnarray}

\subsection{Student Global Latent Factor}
Eq. \ref{eq:EnMF} captures student knowledge factors per term and captures the sequential dynamics in 
student's knowledge state over terms. This can be considered as a latent factor model localized by term. 
%
We propose to incorporate a term-agnostic global latent factor that captures the student-course performance
interaction.  We introduce an  additional 
latent factor $\mathbf{p}_{g(s)}$  that captures the student's implicit 
information (e.g.,  student interest and subject matter mastery toward each knowledge component)  in a common 
latent space as course knowledge components.  The estimated grade of student $s$ on course $c$ at
term $t$ with this global latent factor is given as:
\begin{eqnarray}
\label{eq:ACE}
\begin{aligned}
\tilde g_{s,c}^t  =  {\mathbf{p}_{en(s)}^t}^{\mathsf{T}} \mathbf{q}_{en(c)} + \mathbf{p}_{g(s)}^{\mathsf{T}} \mathbf{q}_c
\end{aligned}
\end{eqnarray}

Here, we compute the dot  product of 
$\mathbf{p}_{g(s)}$ and $\mathbf{q}_c$ instead 
of $\mathbf{p}_{g(s)}$ and $\mathbf{q}_{en(c)}$ in this step. 
The exclusive $l_1$ norm for $\mathbf{p}_{g(s)}$ controls its sparsity since we assume  most students have a tendency to perform well in a fraction of 
the represented knowledge states. 
We refer to this model as \textbf{A}dditive \textbf{L}atent \textbf{E}ffect (ALE).

\subsection{Student and Course Bias Effect}

Inspired by the success of \MF methods  with bias terms \cite{Koren2009}, we add 
student-specific  and course-specific
bias terms denoted by $b_s$ and $b_c$ 
within the \AvgCKRM and \EnFARMa{} formulation in Eq \ref{eq:AvgCKRM} and \ref{eq:ACE}, respectively,   
 as follows: 
\begin{eqnarray}
  \label{eq:AvgCKRMbias}
  \begin{aligned}
	\tilde g_{s,c}^t   = \frac{1}{|G_s^{t-1}|} {\mathbf{p}_{ck(s)}^t}^{\mathsf{T}}  \mathbf{q}_c +  b_s +  b_c
  \end{aligned}
\end{eqnarray}
and
\begin{eqnarray}
  \label{eq:ACEbias}
  \begin{aligned}
	\tilde g_{s,c}^t   = {\mathbf{p}_{en(s)}^t}^{\mathsf{T}} \mathbf{q}_{en(c)} + \mathbf{p}_{g(s)}^{\mathsf{T}} \mathbf{q}_c + b_s + b_c.
  \end{aligned}
\end{eqnarray}
We denote the \AvgCKRM with bias terms as   \AvgCKRMb, and 
\EnFARMa with bias terms as  \EnFARMab. 

Table \ref{tab:methods} summarizes the proposed methods and comparative baselines in terms of their key features and effect-components considered in this study.

\begin{table}
\caption{Method Summarization \label{tab:methods}}
\centering
\begin{threeparttable}
\begin{tabular}{
                @{\hspace{0pt}}>{\small}l@{\hspace{2pt}}
                @{\hspace{2pt}}>{\small}l@{\hspace{2pt}}
                @{\hspace{2pt}}>{\small}l@{\hspace{5pt}}
                @{\hspace{2pt}}>{\small}c@{\hspace{1pt}}
                @{\hspace{2pt}}>{\small}c@{\hspace{1pt}}
                @{\hspace{2pt}}>{\small}c@{\hspace{1pt}}
                @{\hspace{2pt}}>{\small}c@{\hspace{0pt}}
}
\hline\hline
& \multirow{2}{*}{\footnotesize{Method}}  &   \multirow{2}{*}{\footnotesize{Prediction Formulation}}  &  \multicolumn{4}{>{\small}c}{\footnotesize{Property}}  \\
		&	&	 &  $ck$  &  $b_s$  &  $b_c$ &  \tiny{$al, in, g$} \\
	\hline

\multirow{5}{*}{\footnotesize{Baselines}}	&	\MF			&$\mathbf{p}_s^{\mathsf{T}} \mathbf{q}_c$   (Eq. \ref{eq:mf0}) &  $\tickNo$  &  $\tickNo$  &  $\tickNo$  &  $\tickNo$ \\
	
	&	\MFwithbias	&$\mathbf{p}_s^{\mathsf{T}} \mathbf{q}_c + b_s+ b_c$  (Eq. \ref{eq:mf}) & \tickNo &   $\checkmark$  &  $\checkmark$  &  $\tickNo$ \\

	&	\DomainAware	&$\mathbf{p}_s^{\mathsf{T}} \mathbf{q}_c + b_s^{\varphi(c)} + b_c^{\varphi(s)}$  (Eq. \ref{eq:mfd})  &  $\tickNo$  &  $\checkmark$  &  $\checkmark$  &  $\tickNo$  \\

	&	\AvgCKRM{}	&$  \frac{1}{|G_s^{t-1}|} {\mathbf{p}_{ck(s)}^t}^{\mathsf{T}} \cdot \mathbf{q}_c$   (Eq. \ref{eq:AvgCKRM})   &  $\checkmark$  &  $\tickNo$  &  $\tickNo$  &  $\tickNo$ \\

	&	\AvgCKRMb{} &$  \frac{1}{|G_s^{t-1}|}  {\mathbf{p}_{ck(s)}^t}^{\mathsf{T}} \cdot \mathbf{q}_c + b_s+ b_c$   (Eq. \ref{eq:AvgCKRMbias})  &  $\checkmark$  &   $\checkmark$  &  $\checkmark$  &  $\tickNo$  \\

\hline
\footnotesize{Proposed}	&	\EnFARMa	&$(\mathbf{p}_{ck(s)}^t+\mathbf{p}_{al(s)})^{\mathsf{T}} (\mathbf{q}_c+\mathbf{q}_{in(c)})$  &  $\checkmark$  &  $\tickNo$  &  $\tickNo$  &  $\checkmark$ \\
\footnotesize{Methods}	&		&	$ +\mathbf{p}_{g(s)}^{\mathsf{T}} \mathbf{q}_c$    (Eq. \ref{eq:ACE})  &    &   &  &  \\	
	
	&	\EnFARMab	&\EnFARMa $+ b_s + b_c $   (Eq. \ref{eq:ACEbias})   &  $\checkmark$  &  $\checkmark$  &  $\checkmark$  &  $\checkmark$\\	
			
	\hline\hline
\end{tabular}

\begin{tablenotes}[para,flushleft]
\footnotesize{ ``$\checkmark$" indicates the method contains the corresponding property, and ``$\tickNo$" indicates the opposite. 
$al, in, g$ indicate student academic level, course instructor and student global latent factor, respectively. 
$b_s$  and  $b_c$ denote student and course bias terms.
	}
\end{tablenotes} 
\end{threeparttable}

\end{table}

\subsection{Optimization for \EnFARMa}
The optimization problem for \EnFARMa can be formulated as follows:
\begin{eqnarray}
  \label{eq:sitSGPcopt}
  \begin{aligned}
    \underset{\Theta}{\min} \  L(\Theta) + R(\Theta),
  \end{aligned}
\end{eqnarray}
where $\Theta$ represents model parameters (i.e., the latent factors),
$L(\Theta)$ is the loss function and $R(\Theta)$ is the regularization function.
We use a squared error loss function in \EnFARMa:
%
\begin{eqnarray}
  \label{eq:sitSGPcopt}
  \begin{aligned}
    L(\Theta) = \sum_{g_{s,c}^t  \in G_s^{t-1}} (g_{s,c}^t - \tilde g_{s, c}^t(\Theta))^2
  \end{aligned}
\end{eqnarray}
  The $R(\Theta)$ is defined
  as follows:

\begin{eqnarray}
  \label{eq:sitSGPcopt}
  \begin{aligned}
    R(\Theta) =  \sum_{g_{s,c}^t \in G^{T-1}}   &  \bigg[ \frac{\gamma}{2}  ( || {\mathbf{k}_c} || _2 + || {\mathbf{q}_c} || _2 \\
    													& + || {\mathbf{p}_{al(s)}} ||_2  +  || {\mathbf{q}_{in(c)}}  ||_2 +  || {\mathbf{p}_{g(s)}}  ||_2)  \\
    													 & + \alpha_1 (|| {\mathbf{p}_{al(s)}} ||_1 + || {\mathbf{q}_{in(c)}} ||_1)+\alpha_2  || {\mathbf{p}_{g(s)}}  ||_1  \bigg]
  \end{aligned}
\end{eqnarray}

We use stochastic gradient descend (SGD) to solve the
  optimization problem.   The optimization algorithm
  is presented in Algorithm {\ref{alg:sgdsitSGPcopt}}.
\makeatletter
\algnewcommand{\LineComment}[1]{\Statex \hskip\ALG@thistlm #1}
\makeatother

\begin{algorithm} \small\baselineskip=9pt
\caption{\textit{ \text{\EnFARMa{}}}: Learn}\label{alg:sgdsitSGPcopt}
\begin{algorithmic}[1]
\Procedure{\EnFARMa{}\_Learn}{} \label{alg:line1}
\State Initialize $\mathbf{k}_c$, $\mathbf{q}_c$ for each $c$, $\mathbf{p}_{g}$ for each student, $\mathbf{p}_{al}$ for each academic level and $\mathbf{q}_{in}$ for each instructor with random values in (0, 1)
\State $\textit{$\eta$} \gets \text{learning rate}$  \label{alg:line3}
\State $\textit{$\gamma$,$\alpha_1$, $\alpha_2$} \gets \text{regularization weight}$  \label{alg:line4}
\State $\textit{iter} \gets \text{0}$  \label{alg:line5}
\While{iter$<$maxIter and MAE decreases}  \label{alg:line6}
  \For {\textbf{all} $g_{s,c}^t$ $\in$ G$_s^{t-1}$}  \label{alg:line7}
       \State $\mathbf{p}_{ck(s)} \gets 0$  \label{alg:line8}
		 \For {\textbf{all} $c' \in C_s$}   \label{alg:line9}
       	    		\State $\mathbf{p}_{ck(s)}  \gets \mathbf{p}_{ck(s)}  + e^{-\lambda(t_{s,c}-t_{s,c'})} \mathbf{k}_{c'} \cdot g_{s,c'}^{t_{s,c'}}$  \label{alg:line10}
	       	\EndFor
       \State $\tilde g_{s,c}^t\gets (\mathbf{p}_{ck(s)} + \mathbf{p}_{al(s)})^{\mathsf{T}} (\mathbf{q}_c + \mathbf{q}_{in(c)}) +  \mathbf{p}_{g(s)}^{\mathsf{T}} \mathbf{q}_c$   \label{alg:line11}
       \State $e_{s,c}^t = g_{s,c}^t - \tilde g_{s,c}^t$  \label{alg:line12}
       		\For {\textbf{all} $c' \in C_s$} \label{alg:line13}
		\State $\mathbf{k}_{c'} \gets \mathbf{k}_{c'}+$
			\LineComment \hspace*{\algorithmicindent}{\raggedright{$ \eta ( (\mathbf{q}_c + \mathbf{q}_{in(c)}) \cdot e^{-\lambda(t_{s,c}-t_{s,c'})} \cdot  g_{s,c'}^{t_{s,c'}} \cdot e_{s,c}^t- \gamma \cdot \mathbf{k}_{c'})$}}\label{alg:line14} 
		\EndFor	
       \State $\mathbf{p}_{al(s)} \gets \mathbf{p}_{al(s)} + \eta (  (\mathbf{q}_c + \mathbf{q}_{in(c)}) \cdot e_{s,c}^t - \gamma \cdot \mathbf{p}_{al(s)} - \alpha_1)$\label{alg:line15}
        \State $\mathbf{q}_c \gets \mathbf{q}_c + \eta ( (  (\mathbf{p}_{ck(s)} + \mathbf{p}_{al(s)}) +   \mathbf{p}_{g(s)}) \cdot e_{s,c}^t - \gamma \cdot \mathbf{q}_c)$\label{alg:line16}
       \State $\mathbf{q}_{in(c)} \gets \mathbf{q}_{in(c)} +$
      	 \LineComment \hspace*{\algorithmicindent}{\raggedright{$\eta (  (\mathbf{p}_{ck(s)} +  \mathbf{p}_{al(s)}) \cdot e_{s,c}^t - \gamma \cdot \mathbf{q}_{in(c)} - \alpha_1)$}} \label{alg:line17}
        \State $\mathbf{p}_{g(s)} \gets \mathbf{p}_{g(s)} + \eta (  \mathbf{q}_c \cdot e_{s,c}^t - \gamma \cdot \mathbf{p}_{g(s)} - \alpha_2)$	 \label{alg:line18}
  \EndFor
\State $\textit{iter} \gets \textit{iter}+1$ \label{alg:line19}
\EndWhile \\
\Return $\mathbf{k}_c$, $\mathbf{q}_c$, $\mathbf{p}_{g}$, $\mathbf{p}_{al}$  and $\mathbf{q}_{in}$ \label{alg:line20}
\EndProcedure
\end{algorithmic}
\end{algorithm}

\subsection{Computational Complexity Analysis}
The computational complexity of \EnFARMa is determined by the steps from line \ref{alg:line6}
to  line \ref{alg:line19}  in Algorithm \ref{alg:sgdsitSGPcopt}.
In detail, the computational complexity for line \ref{alg:line9} and line \ref{alg:line10}  is upper-bounded by $O(m_c \times k)$, where $m_c$ is the maximum number of courses that a student can take in college. 
For line \ref{alg:line11} and line \ref{alg:line12}, the computational complexity is $O(k^2)$. 
Line \ref{alg:line13} and line \ref{alg:line14} have complexity $O(m_c \times k)$ as well. 
From line \ref{alg:line15} to line \ref{alg:line18}, the total computational complexity is $O(4 \times k)$. 
Thus, 
the computational complexity for \EnFARMa is $O(n_{iter} \times n_g \times (2m_c \times k + k^2 + 4 \times k))$, where $n_{iter}$ is the
  number of iterations, $n_g$ is the total number of student-course grades, $m_c$ is the maximum number
  of courses that a student can take, and $k$ is the dimension of latent factors. Typically, $m_c > k$ and thus the complexity is $O(n_{iter} \times n_g \times m_c \times k)$.

\section{Experiments}
\subsection{Dataset Description}
We evaluated our methods on student grade 
data obtained from 
George Mason University. The data was extracted in 
the period of Fall 2009 to Spring 2016 and includes information
for 23,013 transfer students (TR) and 
20,086 first-time freshmen (FTF; i.e., students who 
begin their study at  George Mason University) 
across 151 majors enrolled in 4,654 courses for both TR and FTF students.

Specifically, we evaluated the proposed models 
on 
nine large and diverse majors including: 
(i) Mathematical Sciences (MATH), (ii) Physics (PHYS), (iii) Chemistry (CHEM) (iv) Computer Science (CS), (v) Civil, Environmental and Infrastructure 
Engineering (CEIE), (vi)  Biology (BIOL), (vii) Psychology (PSYC), and (viii) Applied Information Technology  (AIT).
Table  \ref{tab:statics} presents the
details 
about  these 
majors. 

\begin{table}
\caption{Dataset Statistics}~\label{tab:statics}
\centering
\begin{small}
\begin{threeparttable}
\begin{tabular}{r| r r r | r r r}
\hline\hline
 \multirow{2}{*}{Major} & \multicolumn{3}{ c|}{FTF Students} & \multicolumn{3}{c}{TR Students} \\
\cline{2-7}
& \#S &  \#C & \#S-C & \#S & \#C &  \#S-C\\
\hline
MATH 	&   209	&   84 	&    	2,846	&258  	& 91 		&   2,580	\\

PHYS 	& 127 	& 53 		&    	1,830	&74  		& 48 		&   854	\\

CHEM 	& 342 	&  55 	&    	4,649	& 278 	& 66 		& 3,105	\\

CS   		& 988  	&  76  	& 	13,809	& 554 	&68  		& 7,028	\\

CEIE 	&  428	&  80 	&    6,925		& 248 	& 92 		&4,036	\\

BIOL 	& 1,629 	&  109 	&    21,519	& 1,525 	& 115 	& 16,615	\\

PSYC 	& 1,114 	& 95  	&    	14,377	& 1,749 	& 114 	& 18,939	\\

AIT 		& 334 	& 82 		&   6088		& 1,170 	&90 		&15,060	\\

\hline

Total 	& 5,171 	& 634  	&  72,043  	& 5,856  	& 684 	&  68,216	 \\

\hline\hline
\end{tabular}

\begin{tablenotes}[para,flushleft]
\footnotesize{\#S, \#C and \#S-C are  number of students, courses and student-course grades from Fall 2009 to Spring 2016, respectively.}
\end{tablenotes}

\end{threeparttable}
\end{small}
\end{table}

\subsection{Experimental Protocols}

To assess the performance of the next-term
grade prediction models, we train our models on data up to term $T-1$ and make predictions 
for term $T$. We evaluate our methods for three 
  test terms, i.e., Spring 2016, Fall 2015 and Spring 2015.  As an 
  example, to evaluate predictions for term 
  Fall 2015, data from Fall 2009 to Spring 2015 are  
 used for model training and data from Fall 2015 are used as the 
 test data. 

\subsection{Parameter Learning}
\label{ParameterLearning}
The parameters in the optimization problem (Eq \ref{eq:sitSGPcopt}) contain the number of latent dimensions (i.e., $k$), regularization weights (i.e., $\gamma$, $\alpha_1$, $\alpha_2$), and time decay parameter (i.e., $\lambda$). We use a validation set to select parameters. 
Specifically, for test term $T$, we have student-course grades up to term $T-1$ as the training set, i.e., $G^{T-1}$.  Then we split the training set into two parts:   $G^{T-2}$ and $G_{T-1}$, the latter of which we consider as the validation set. 
We did a grid search over the parameters and selected the parameters that perform best on the validation set.

\subsection{Evaluation Metrics}
In our dataset, a student's grade is a letter grade  (i.e.  A, A-, \ldots, F).
As done previously 
in Polyzou \emph{et. al.} \cite{polyzou2016grade},  we compute the 
\textbf{P}ercentage of \textbf{T}ick \textbf{A}ccuracy (PTA).
First, we define
a tick as the difference between two consecutive letter grades (e.g., C+ vs C or C vs C-). To assess
the performance of our grade prediction methods, we convert
the predicted numerical grades into their closest letter grades.
Specifically, we set letter grade ``A+" and ``A" correspond to number 4.0, ``A-" correspond to number 3.67, ``B+" correspond to number 3.33, etc.  
In our experiments, we first convert letter grades to numbers during training, and then convert the predicted numbers to letter grades during  testing 
and
 compute the percentage of 
predicted grades with no error (or 0-ticks), within 1 tick and within 2 ticks denoted
by \PTAzero{}, \PTAone{} and \PTAtwo{}, respectively. For 
course selection and degree 
planning purposes, courses predicted within 2 ticks can be considered sufficiently close.

We use
Mean Absolute Error (MAE)  
for evaluating the predicted results with numbers. MAE is given as:
\begin{equation}
\begin{aligned}
\begin{split}
MAE = \frac{\sum_{s,c\in G_T}\left |g_{s,c} - \tilde{g}_{s,c} \right |}{\left | G_T  \right |}
\end{split}
\end{aligned}
\end{equation}
where $g_{s,c}$ and $\tilde{g}_{s,c}$ are the ground truth and predicted grade for student $s$ on course $c$, respectively.  $G_T$ is the test 
set of (student, course, grade) triples in the $T$-th term.

\begin{table*}[!h]
\caption{Performance Comparison for All Methods}~\label{tab:general}
\centering
\begin{tabular}{
                @{\hspace{3pt}}>{\small}r@{\hspace{8pt}}
                @{\hspace{6pt}}>{\small}r@{\hspace{6pt}}
                @{\hspace{6pt}}>{\small}r@{\hspace{6pt}}
                @{\hspace{6pt}}>{\small}r@{\hspace{6pt}}
                @{\hspace{6pt}}>{\small}r@{\hspace{3pt}}
                @{\hspace{8pt}}>{\small}r@{\hspace{8pt}}
                @{\hspace{6pt}}>{\small}r@{\hspace{6pt}}
                @{\hspace{6pt}}>{\small}r@{\hspace{6pt}}
                @{\hspace{6pt}}>{\small}r@{\hspace{6pt}}
                @{\hspace{6pt}}>{\small}r@{\hspace{6pt}}
                @{\hspace{6pt}}>{\small}r@{\hspace{3pt}}
                @{\hspace{8pt}}>{\small}r@{\hspace{8pt}}
                @{\hspace{6pt}}>{\small}r@{\hspace{6pt}}
                @{\hspace{6pt}}>{\small}r@{\hspace{6pt}}
                @{\hspace{6pt}}>{\small}r@{\hspace{6pt}}
                @{\hspace{6pt}}>{\small}r@{\hspace{6pt}}
                @{\hspace{6pt}}>{\small}r@{\hspace{6pt}}
}
  \hline\hline
 \multirow{2}{*}{Method}  &   \multicolumn{7}{>{\small}c}{FTF - Spring 2016} & 	&\multicolumn{7}{>{\small}c}{TR - Spring 2016}   \\
\cline{2-8}
\cline{10-16}
&  \multicolumn{3}{>{\small}c}{parameters} & MAE & \PTAzero & \PTAone  & \PTAtwo & 	& \multicolumn{3}{>{\small}c}{parameters} & MAE & \PTAzero & \PTAone  & \PTAtwo \\
\hlineB{1}
\MFnobias 		& 	10 & -- & --	&0.723    & 0.188	&0.338 &  0.580 &      &	10 & -- & --	 &  0.706  &  0.188  &  0.341  &  0.601	\\
\MFwithbias 	 & 10 & -- & -- 	 &  0.670  &  0.206  &  0.360  &  0.609 &      &	10 & -- & --	&  0.658  &  0.226  &  0.387  &  0.628	\\
\DomainAware  & 5 & -- & --  	&  0.661   & 0.221  &  0.381  &  0.621 &      &	10 & -- & --	 &  0.683  &  0.216  &  0.366  &  0.614	\\
\AvgCKRM	 & 5 & 0.01 & --	  &  0.674  &  0.216  &  0.362  &  0.604 &      &	5 & 0.01 & --	  &  0.680  &  0.225  &  0.369  &  0.597	\\
\AvgCKRMb	 & 5 & 0.01 & --	  &  0.647 & 0.218&	0.379&	0.625&      & 5 & 0.01 & --	  &   0.658 & 0.227 & 0.387 & 0.627\\
\hline
\EnFARMa  & 0.01		&0.01 & 0.1	& {\bf 0.625}&{\bf 0.255}&{\bf0.416}&{\bf0.651}&	&0.01 & 0.001&	0.1  &  0.645&  {\bf 0.247}  &   {\bf 0.395}  &  {\bf 0.651}	\\
\EnFARMab  & 0.1		&0.01 & 0.1	&{\bf 0.625}    &	0.225    &	0.389     &	0.648 &      &	0.01	&0.1 & 0.01	& {\bf 0.642} & 0.231 & 0.389 & 0.637\\
\hline \hline
\end{tabular}

\begin{tabular}{
                @{\hspace{3pt}}>{\small}r@{\hspace{8pt}}
                @{\hspace{6pt}}>{\small}r@{\hspace{6pt}}
                @{\hspace{6pt}}>{\small}r@{\hspace{6pt}}
                @{\hspace{6pt}}>{\small}r@{\hspace{6pt}}
                @{\hspace{6pt}}>{\small}r@{\hspace{6pt}}
                @{\hspace{6pt}}>{\small}r@{\hspace{6pt}}
                @{\hspace{6pt}}>{\small}r@{\hspace{6pt}}
                @{\hspace{6pt}}>{\small}r@{\hspace{6pt}}
                @{\hspace{6pt}}>{\small}r@{\hspace{6pt}}
                @{\hspace{6pt}}>{\small}r@{\hspace{6pt}}
                @{\hspace{6pt}}>{\small}r@{\hspace{6pt}}
                @{\hspace{6pt}}>{\small}r@{\hspace{6pt}}
                @{\hspace{6pt}}>{\small}r@{\hspace{6pt}}
                @{\hspace{6pt}}>{\small}r@{\hspace{6pt}}
                @{\hspace{6pt}}>{\small}r@{\hspace{6pt}}
                @{\hspace{6pt}}>{\small}r@{\hspace{6pt}}
                @{\hspace{6pt}}>{\small}r@{\hspace{6pt}}
}
 \multirow{2}{*}{Method}  &   \multicolumn{7}{>{\small}c}{FTF - Fall 2015} & 	&\multicolumn{7}{>{\small}c}{TR - Fall 2015}   \\
\cline{2-8}
\cline{10-16}
&  \multicolumn{3}{>{\small}c}{parameters}  & MAE & \PTAzero & \PTAone  & \PTAtwo &  	&\multicolumn{3}{>{\small}c}{parameters}  & MAE & \PTAzero & \PTAone  & \PTAtwo \\
\hlineB{1}
\MFnobias 		& 	10 & -- & --	  &  0.730  &  0.177  &  0.317  &  0.574 &      &	10 & -- & --	  &  0.692  &  0.183  &  0.347  &  0.599	\\
\MFwithbias 	 & 10 & -- & --	  &  0.691  &  0.205  &  0.360  &  0.605 &      &	10 & -- & --	  &  0.653  &  0.213  &  0.378  &  0.631	\\
\DomainAware  & 10 & -- & --	  &  0.693  &  0.216  &  0.370  &  0.610 &      &	10 & -- & --	  &  0.670  &  0.205  &  0.362  &  0.630	\\
\AvgCKRM	 & 5 & 0.01 & --	  &  0.706  &  0.193  &  0.347  &  0.585 &      &	5 & 0.01 & --	  &  0.665  &  0.210  &  0.372  &  0.616	\\
\AvgCKRMb	 & 5 & 0.01 & --	  &  0.690	&0.195 & 0.351&	0.603&      & 5 & 0.01 & --	  &  	0.642 & 0.227 & 0.394 & 0.641	\\
\hline
\EnFARMa &	0.1 & 0.001&	0.05	  & {\bf 0.654}  & {\bf  0.251}  &  {\bf 0.400}  &  {\bf 0.638} &      &	0.01 & 0.001&	0.05	  &  {\bf 0.615} &  {\bf 0.243}  &  {\bf 0.418}  &  {\bf 0.670}	\\
\EnFARMab  & 0.01		&0.01 & 0.1 & 0.660 & 0.223 & 0.379 & 0.634 &      &	0.01	&0.01 & 0.1 & 0.627 & 0.216 & 0.392 & 0.655\\
\hline \hline
\end{tabular}

\begin{threeparttable}

\begin{tabular}{
                @{\hspace{3pt}}>{\small}r@{\hspace{8pt}}
                @{\hspace{6pt}}>{\small}r@{\hspace{6pt}}
                @{\hspace{6pt}}>{\small}r@{\hspace{6pt}}
                @{\hspace{6pt}}>{\small}r@{\hspace{6pt}}
                @{\hspace{6pt}}>{\small}r@{\hspace{6pt}}
                @{\hspace{6pt}}>{\small}r@{\hspace{6pt}}
                @{\hspace{6pt}}>{\small}r@{\hspace{6pt}}
                @{\hspace{6pt}}>{\small}r@{\hspace{6pt}}
                @{\hspace{6pt}}>{\small}r@{\hspace{6pt}}
                @{\hspace{6pt}}>{\small}r@{\hspace{6pt}}
                @{\hspace{6pt}}>{\small}r@{\hspace{6pt}}
                @{\hspace{6pt}}>{\small}r@{\hspace{6pt}}
                @{\hspace{6pt}}>{\small}r@{\hspace{6pt}}
                @{\hspace{6pt}}>{\small}r@{\hspace{6pt}}
                @{\hspace{6pt}}>{\small}r@{\hspace{6pt}}
                @{\hspace{6pt}}>{\small}r@{\hspace{6pt}}
                @{\hspace{6pt}}>{\small}r@{\hspace{6pt}}
}
 \multirow{2}{*}{Method}  &   \multicolumn{7}{>{\small}c}{FTF - Spring 2015} & 	&\multicolumn{7}{>{\small}c}{TR - Spring 2015}   \\
\cline{2-8}
\cline{10-16}
&  \multicolumn{3}{>{\small}c}{parameters}  & MAE & \PTAzero & \PTAone  & \PTAtwo &  	&\multicolumn{3}{>{\small}c}{parameters}  & MAE & \PTAzero & \PTAone  & \PTAtwo \\
\hlineB{1}
\MFnobias 		& 	10 & -- & --	  &  0.760  &  0.168  &  0.306  &  0.547 &      &	10 & -- & --	  &  0.743  &  0.169  &  0.316  &  0.559	\\
\MFwithbias 	 & 10 & -- & --	  &  0.718  &  0.186  &  0.335  &  0.582 &      &	10 & -- & --	  &  0.688  &  0.218  &  0.368  &  0.607	\\
\DomainAware  & 10 & -- & --	  &  0.716  &  0.215  &  0.358  &  0.595 &      &	10 & -- & --	  &  0.693  &  0.229  &  0.383  &  0.618	\\
\AvgCKRM	 & 5 & 0.01 & 0.01	  &  0.712  &  0.192  &  0.332  &  0.579 &      &	5 & 0.01 & 0.01	  &  0.705  &  0.214  &  0.357  &  0.589	\\
\AvgCKRMb	 & 5 & 0.01 & 0.01	  &  0.690 & 0.203&	0.354&	0.599&      & 5 & 0.01 & 0.01	  & 	0.688 & 0.207 & 0.354 & 0.606	\\
\hline
\EnFARMa &	0.01 & 0.001&	0.1	  &{\bf 0.649}&{\bf 0.244}&{\bf 0.403}&{\bf 0.639}&      & 0.01 & 0.001&	0.1  & {\bf 0.644}&  {\bf 0.254}  &{\bf 0.417}  & {\bf 0.647}\\
\EnFARMab  & 0.01		&0.01 & 0.1 & 0.657 & 0.214 & 0.372 & 0.618 &      &	0.1	&0.01 & 0.01 & 0.653 & 0.226 & 0.383 & 0.631\\
\hline\hline
\end{tabular}

\begin{tablenotes}[para,flushleft]
\footnotesize{ Columns under ``parameters" indicate different model parameters for the corresponding methods. Specifically, for \MFnobias, \MFwithbias and \DomainAware, the parameter is the dimension of latent factors, ($k$). 
For \AvgCKRM and \AvgCKRMb, 
the parameters are the dimension of latent factors, ($k$), and
time-decay coefficient, ($\lambda$). 
For \EnFARMa and \EnFARMab, the parameters are  time-decay coefficient, ($\lambda$), regularization weight for $\mathbf{p}_{al(s)}$ and $\mathbf{q}_{in(c)}^t$, ($\alpha_1$), 
and regularization weight for $\mathbf{p}_{g(s)}$, ($\alpha_2$). Bold numbers are the best performing results. In our experiments, we select the best performance for all baseline methods.}
\end{tablenotes} 
\medskip
\end{threeparttable} 
\end{table*}

\section{Results and Discussion}

\subsection{Overall Performance}

Table \ref{tab:general} shows the comparison of MAE
and PTA results  for FTF and TR students across 
Spring 2016, Fall 2015 and Spring 2015 test terms.  
The parameters are determined as discussed in Section \ref{ParameterLearning}.  
In our reported results, 
\DomainAware \cite{elbadrawy2016domain} 
has nine different combinations for student- and course-level groupings 
and has bias.  We  tried all the proposed combinations 
and 
report
the best performance among all the results.  
The results 
show that \EnFARMa has the best performance 
on all the evaluation metrics (the only exception is in Spring 2016 on MAE). 
%
%
Specifically, \EnFARMa outperforms the baseline methods 
on {\PTAzero}, {\PTAone} and {\PTAtwo}  by 10.61\%, 7.17\% and 4.50\%, respectively. 
We also observe that the improvement in performance of \EnFARMa over the baseline approaches is greater 
for Spring 2015 in comparison to Spring 2016, even though the training set for Spring 2015 is smaller than Spring 2016. 
This shows that \EnFARMa can 
overcome the scarcity issues in a dataset and yield good prediction performance.

\subsection{Effects of Bias Terms}
For all the datasets in Table \ref{tab:general},
\MFwithbias (i.e., \MF with student/course-specific bias terms)
and
\DomainAware (i.e., \MF with domain-aware biases)
always outperform \MFnobias  (i.e., \MF without bias terms).
In addition, \MFwithbias  achieves
better \PTAzero
 on 
 TR students,  
  but worse \PTAzero on FTF students than \DomainAware.
  This is probably because the FTF students show consistent characteristics in comparison to TR students, who typically have more diverse backgrounds.

In Table \ref{tab:general}, we also observe that \AvgCKRMb 
consistently outperforms \AvgCKRM, similar to the comparison between \MFwithbias and \MFnobias, 
but  \EnFARMa always outperforms \EnFARMab.   
This may indicate that the additive latent effects in \EnFARMa have also captured the student and course bias information in \EnFARMab. 
The results in Table \ref{tab:general} demonstrate that \EnFARMa is able to achieve better prediction performance without explicitly modeling student and course biases.

\begin{figure*}[h!]
  \centering
  \includegraphics[width=1\linewidth]{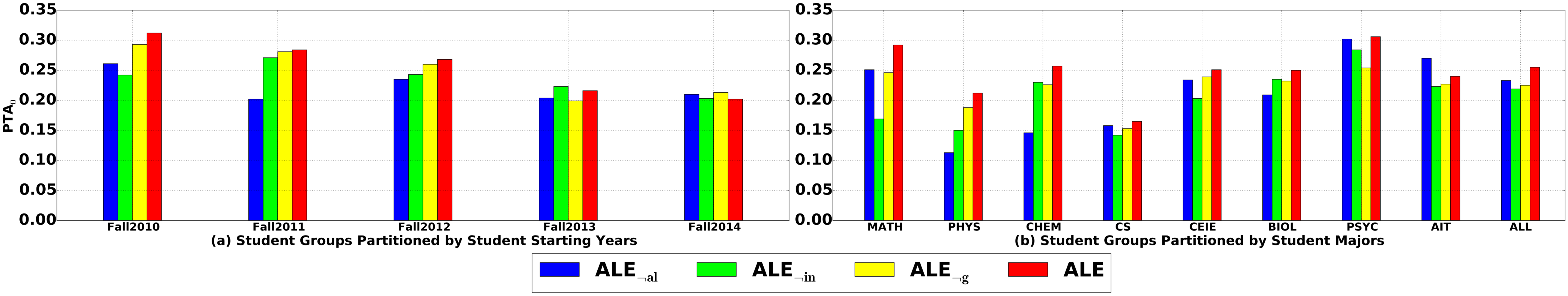}
  \caption{Comparison of \PTAzero with Each Effect Removed on Various Student Groups in \EnFARMa}
  \label{fig:ntr-factors}
 \end{figure*}

\subsection{Importance of Additive Latent Effects}
In order to learn the importance of each additive latent effect, we perform a 
study to assess the prediction performance  
of different \EnFARMa models with a particular latent effect removed. Table \ref{tab:importance} shows the details of the compared models in this experiment.   
Fig. \ref{fig:ntr-factors} represents the \PTAzero performance of each \EnFARMa model variant.  
We test the results  on various student groups partitioned by student starting years and student majors, as shown in Fig. \ref{fig:ntr-factors}a and Fig. \ref{fig:ntr-factors}b, respectively.  
 We also implement the experiment for the whole test set, i.e. $G_T$, and present the results with label ``ALL" in Fig. \ref{fig:ntr-factors}b.  

\begin{table}
\caption{Comparison Method Summarization \label{tab:importance}}
\centering
\begin{threeparttable}
\begin{tabular}{
                @{\hspace{3pt}}>{\footnotesize}r@{\hspace{3pt}}
                @{\hspace{5pt}}>{\small}l@{\hspace{30pt}}
                @{\hspace{2pt}}>{\small}l@{\hspace{2pt}}
                @{\hspace{5pt}}>{\small}l@{\hspace{5pt}}
                @{\hspace{3pt}}>{\small}c@{\hspace{3pt}}
}
\hline\hline

 \multirow{2}{*}{\footnotesize{Method}}  &   \multirow{2}{*}{\footnotesize{Prediction Formulation}}  &  \multicolumn{3}{>{\small}c}{\footnotesize{Property}}  \\
		&	&  $al$ &  $in$  &  $g$ \\
	\hline


\EnFARMaal			 	&	   ${\mathbf{p}_{ck(s)}^t}^{\mathsf{T}}  (\mathbf{q}_c+\mathbf{q}_{in(c)})+\mathbf{p}_{g(s)}^{\mathsf{T}} \mathbf{q}_c$ &  $\tickNo$  &  $\checkmark$&  $\checkmark$\\

\EnFARMain			 	&	    $(\mathbf{p}_{ck(s)}^t+\mathbf{p}_{al(s)})^{\mathsf{T}}  \mathbf{q}_c+\mathbf{p}_{g(s)}^{\mathsf{T}}\mathbf{q}_c$ &  $\checkmark$  &  $\tickNo$&  $\checkmark$\\
	
\EnFARMaI				&	    $(\mathbf{p}_{ck(s)}^t+\mathbf{p}_{al(s)})^{\mathsf{T}}  (\mathbf{q}_c+\mathbf{q}_{in(c)})$ &  $\checkmark$  &  $\checkmark$&  $\tickNo$ \\

\EnFARMa			&$(\mathbf{p}_{ck(s)}^t+\mathbf{p}_{al(s)})^{\mathsf{T}} (\mathbf{q}_c+\mathbf{q}_{in(c)})$  &  $\checkmark$  &  $\checkmark$  &  $\checkmark$ \\
						&	$ +\mathbf{p}_{g(s)}^{\mathsf{T}} \mathbf{q}_c$    (Eq. \ref{eq:ACE})  &    &   &    \\
	\hline\hline
\end{tabular}

\begin{tablenotes}[para,flushleft]
\footnotesize{$al$, $in$  and $g$ indicate the property of student academic level, course instructor and student global latent factor, respectively. $\checkmark$ indicates the model contains the corresponding property, $\tickNo$ indicates the opposite.
}
\end{tablenotes} 
\end{threeparttable}

\end{table}

%
%
Fig. \ref{fig:ntr-factors} shows that for most student groups, \EnFARMa outperforms the other models, indicating that each additive latent effect plays an important role in \EnFARMa. 
%
%
Specifically, Fig. \ref{fig:ntr-factors}a shows that for students who start school in 
Fall 2011, the \PTAzero of \EnFARMaal (without the academic level) 
drops the most 
compared 
to other models. This shows that the 
student  academic level is the most important effect for this student group. Moreover, for students who start school in 
Fall 2013 and Fall 2014, \EnFARMa does not 
outperform all the other models. This indicates that for these two student  groups, it is 
not  necessary to consider all the latent effects when predicting their grades.

%
From Fig. \ref{fig:ntr-factors}b,  we notice that for students in MATH, CS and AIT major, \EnFARMain has the worst \PTAzero results, indicating the course instructor associated latent factor is 
the most important for grade prediction. While, for students  in PHYS, CHEM and BIOL majors,  student academic level  is the most important effect. 
Moreover,  Fig. \ref{fig:ntr-factors}b also shows that for all the students (``ALL"),  course instructor 
and student global latent factor are  more important than student academic level effect. The  \EnFARMa outperforms the other three variants.
For students with different majors and academic levels, all three effects are important in providing an accurate grade prediction.



 \begin{figure}[!h]
    \centering
    \begin{subfigure}[b]{0.5\textwidth}
        \centering
        \includegraphics[height=1.6in]{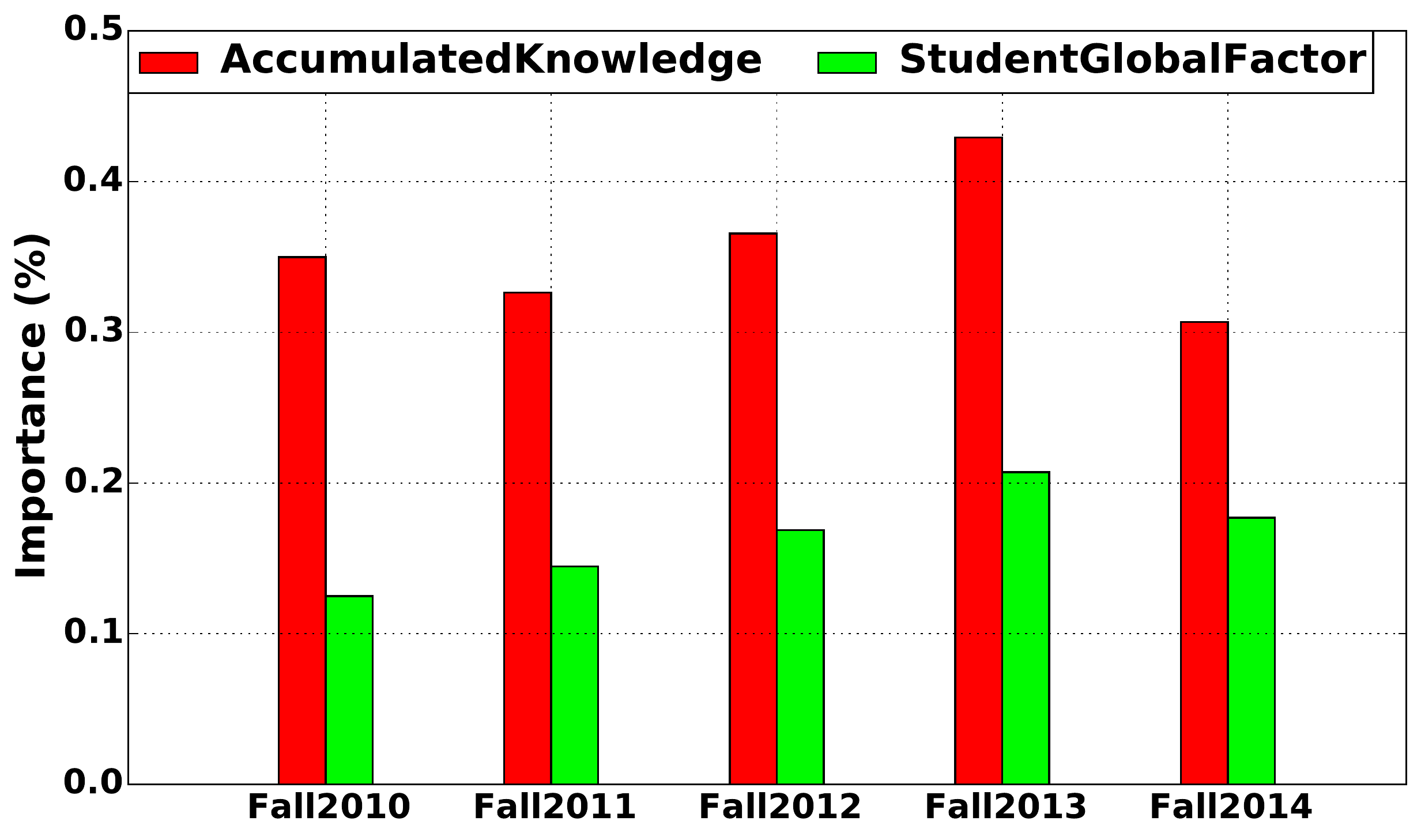}
        \caption{Student Groups Partitioned by Student Starting Years}~\label{fig:importance-cohort}
    \end{subfigure}%
    \hspace{0.005\textwidth}
    \begin{subfigure}[b]{0.5\textwidth}
        \centering
        \includegraphics[height=1.3in]{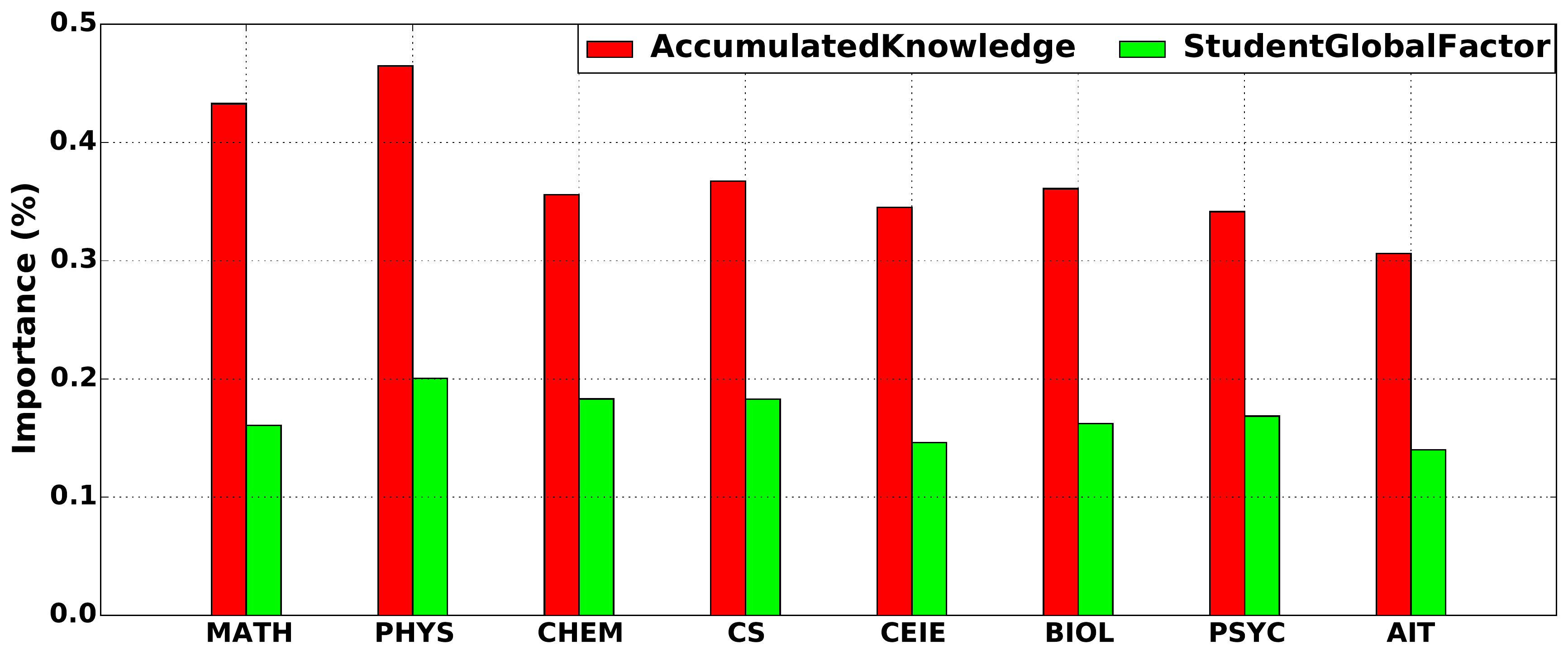}
        \caption{Student Groups Partitioned by Student Majors}~\label{fig:importance-major}
    \end{subfigure}
    \caption{The Importance of Student's Accumulated Knowledge and Student Global Latent Factor on Various Student Groups}
    \label{fig:importance}
\end{figure}

\subsection{Importance of Accumulated Knowledge and Student Global Latent Factor}
Students need help in course selections both in order to gain course credits and learn the knowledge and skills contained in the course.  
In \EnFARMa, accumulated knowledge and student global latent factor are the two effects that are directly related to the students. 
Learning the importance of these two factors 
can assist students when they choose a course.

Specifically, we calculate the importance of each factor 
by averaging the proportion of its contribution in all the predicted grades within the test set 
as follows:
 \begin{eqnarray}
   \label{eq:impo01}
   \begin{aligned}
    I_{ck}  = \frac{1}{|G_T|} \sum_{g_{s,c}^T \in G_T} \frac{\mathbf{p}_{ck(s)}^T}{\tilde g_{s,c}^T}
   \end{aligned}
 \end{eqnarray}
and
 \begin{eqnarray}
   \label{eq:impo02}
   \begin{aligned}
     I_{g}  = \frac{1}{|G_T|} \sum_{g_{s,c}^T \in G_T} \frac{\mathbf{p}_{g(s)}^{\mathsf{T}} \mathbf{q}_c}{\tilde g_{s,c}^T}
   \end{aligned}
 \end{eqnarray}
where $I_{ck}$ and $I_{g}$ represent the importance of accumulated knowledge and student global latent factor, respectively.

We present this experiment on  students  partitioned by  starting years and student majors in Fig. \ref{fig:importance}. 
For all student groups,  
 accumulated knowledge is always more important than student global latent factor. 
 Specifically, Fig. \ref{fig:importance-cohort} shows that for students who start school in Fall 2013, the proportion of accumulated knowledge is the highest among all student groups and
 it is the lowest for students who start school in Fall 2014. 
 Moreover, for students who start school in
 Fall 2010, the proportion of  student global latent factor is the lowest among all student groups. We also notice that the difference between the two factors 
 is the smallest for students who start school in 
 Fall 2014.  
 Fig. \ref{fig:importance-major} shows the results for student groups partitioned by student majors.  It shows that  accumulated knowledge is  more important than student global latent factor 
 for  MATH and PHYS majors. 
AIT has the smallest difference between accumulated knowledge and student global latent factor. 

Based on the results of this experiment, students can balance the course knowledge and their own capabilities 
when selecting courses. For example, CS students who start school in Fall 2013 have the reference information that about 40\% and 20\% of their grades are influenced by the accumulated knowledge and 
student global latent factor, respectively.

\section{Conclusion and Future Work}
In this paper, we presented additive latent effect models, which incorporate additive latent effects associated with students and courses to solve the next-term grade prediction problem. Specifically, we were able to 
highlight the improved performance of \EnFARMa with use of latent factors of course instructors, student academic levels and student global latent effect. 
Our experimental results demonstrate that \EnFARMa 
outperforms all the state-of-the-art baselines 
in various experiments. Specifically, \EnFARMa model outperforms 
the best results among baselines for \PTAzero, \PTAone and \PTAtwo by 10.61\%, 7.17\% and 4.50\%, respectively. 
Moreover, we implemented different sets of experiments to analyze the importance of different effects contained 
in \EnFARMa. 

In the future, we plan to add more factors, such as student's interests and diligence, and  
build a degree planner which can directly recommend courses to students based on these factors. 
We hope such a recommender system can not only help students finish their study at college 
but also guide them plan for careers in the future.






\section*{Acknowledgment}
Thanks to NSF Big Data Grant  \#1447489 and GMU IRR Staff for providing data.


\bibliography{mybib}{}
\bibliographystyle{plain}


\end{document}